\documentclass[runningheads]{llncs}
\usepackage{graphicx}
\usepackage{algorithm2e}
\RestyleAlgo{ruled}
\usepackage{breakcites}

\usepackage{tikz}
\usepackage{comment}
\usepackage{amsmath,amssymb} % define this before the line numbering.
\usepackage{color}
\usepackage{booktabs}
\usepackage{multirow}
\usepackage{hyperref}
\usepackage{graphicx, wrapfig, lipsum}
\usepackage{sidecap}
\sidecaptionvpos{figure}{t}
\usepackage[skip=0.1\baselineskip]{caption}

\usepackage{soul}

\usepackage[accsupp]{axessibility}  % Improves PDF readability for those with disabilities.

\begin{document}
% \linenumbers
\pagestyle{headings}
\mainmatter
\def\ECCVSubNumber{7077}  % Insert your submission number here

\title{HM:~Hybrid Masking for Few-Shot Segmentation} % Replace with your title

\begin{comment}
\titlerunning{ECCV-22 submission ID \ECCVSubNumber} 
\authorrunning{ECCV-22 submission ID \ECCVSubNumber} 
\author{Anonymous ECCV submission}
\institute{Paper ID \ECCVSubNumber}
\end{comment}
%******************

%\begin{comment}
\titlerunning{HM:~Hybrid Masking for Few-Shot Segmentation}
\author{Seonghyeon Moon\inst{1} \and
Samuel S. Sohn\inst{1} \and
Honglu Zhou\inst{1} \and
Sejong Yoon\inst{2} \and
Vladimir Pavlovic\inst{1} \and
Muhammad Haris Khan\inst{3} \and
Mubbasir Kapadia\inst{1} 
} 

\authorrunning{S.~Moon et al.}
\institute{Rutgers University, New Jersey, USA
\email{\{sm2062,samuel.sohn,honglu.zhou,vladimir,mubbasir.kapadia\}@rutgers.edu}
\and
The College of New Jersey, New Jersey, USA
\email{yoons@tcnj.edu}\\
\and
Mohamed Bin Zayed University of Artificial Intelligence, Abu Dhabi, UAE\\
\email{muhammad.haris@mbzuai.ac.ae}}
%\end{comment}
%******************
\maketitle

\newcommand\red[1]{\textcolor{red}{#1}}
\newcommand\blue[1]{\textcolor{blue}{#1}}

\newcommand\revis[2]{\red{\st{#1}} \blue{#2}}

\newcommand\train[1]{$#1_{train}$}
\newcommand\test[1]{$#1_{test}$}

\begin{abstract}
We study few-shot semantic segmentation that aims to segment a target object from a query image when provided with a few annotated support images of the target class. 
Several recent methods resort to a feature masking~(FM) technique to discard irrelevant feature activations which eventually facilitates the reliable prediction of segmentation mask. A fundamental limitation of FM is the inability to preserve the fine-grained spatial details that affect the accuracy of segmentation mask, especially for small target objects.
In this paper, we develop a simple, effective, and efficient approach to enhance feature masking~(FM). We dub the enhanced FM as hybrid masking~(HM). Specifically, we compensate for the loss of fine-grained spatial details in FM technique by investigating and leveraging a complementary basic input masking method.
Experiments have been conducted on three publicly available benchmarks with strong few-shot segmentation~(FSS) baselines.
We empirically show improved performance against the current state-of-the-art methods by visible margins across different benchmarks.
Our code and trained models are available at: \url{https://github.com/moonsh/HM-Hybrid-Masking}

% In particular, HMFS delivers an absolute gain of 6.0\% over \cite{zhang2021fewshot} with ResNet-101~\cite{Resnet} backbone in 1-shot setting on the challenging COCO-20$^i$\cite{lin2015microsoft}.
% %
% Furthermore, compared to baseline \cite{HSNet}, it speeds up the training convergence up to 13x times on COCO-20$^i$\cite{lin2015microsoft}.

\end{abstract}

% face the challenge of accurately segmenting target object information only with limited information.

% Few-shot segmentation methods face the difficulty of accurately grasping target object information only with limited information. We investigate the two different masking approaches for constructing features, input masking (IM) and masked average pooling (MAP). MAP has the advantage of allowing the target object and background information to be included in the feature map. On the other hand, IM creates a feature map of the target object excluding background information. We propose hybrid masking for few-shot segmentation(HMFS) that merges the two feature maps generated by the two masking approaches to produce more meaningful feature maps. This simple change maximizes cosine-similarity between query and support features, and this results in surprising improvement with the HSNet framework. The outstanding performance on the most challenging benchmark in few-shot segmentation, COCO-20$^i$, demonstrates the effectiveness of the proposed method. We also verified the robustness of the HMFS using the domain shift test. Furthermore, HMFS helps a network converge very fast up to 13 times compared to using the MAP while performing the state of the art

\section{Introduction}
\label{sec:intro}

Deep convolutional neural networks~(DCNNs) have enabled remarkable progress in various important computer vision~(CV) tasks, such as image recognition~\cite{krizhevsky2012imagenet,Resnet,VGG16,huang2017densely}, object detection~\cite{ren2015faster,redmon2016you,liu2016ssd}, and semantic segmentation~\cite{Deeplab,fully_conv_net_seg,pyramidscene}. Despite proving effective for various CV tasks, DCNNs require a large amount of labeled training data, which is quite cumbersome and costly to acquire for dense prediction tasks, such as semantic segmentation. Furthermore, these models often fail to segment novel~(unseen) objects when provided with very few annotated training images. To counter the aforementioned problems, few shot segmentation~(FSS) methods, that rely on a few annotated support images, have been actively studied~\cite{Co-FCN,AMP-2,PMM,FWB,PANet,CANet,PFENet,DAN,RePRI,SAGNN,FSOT,Zhang2020SGOneSG,CWT,ASGNet,HSNet}.

%Semi-supervised segmentation~\cite{semi_1,semi_2,semi_3} is an alternative to data-hungry limitation, however, it still relies on a lot of weakly-labeled data for training. These models demonstrate poor generalization when facing unseen classes.  

%Semi-supervised segmentation~\cite{semi_1,semi_2,semi_3} is an alternative to this limitation, however, it still relies on a lot of weakly-labeled data for training.

\begin{figure}[t]
\centering
\includegraphics[height=6.7cm]{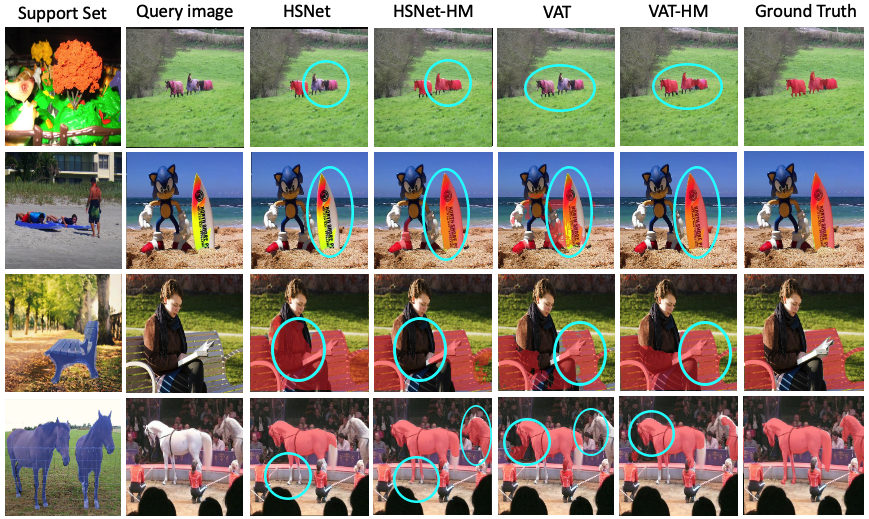}
\caption{ \small Our HM allows generation of more accurate segmentation masks by recovering the fine-grained details~(marked in cyan ellipse) when integrated into the current state-of-the-art methods, HSNet~\cite{HSNet} and VAT~\cite{VAT} on COCO-20$^i$~\cite{lin2015microsoft}.} 

\label{fig:teaser1}
\end{figure}

After the pioneering work of OSLSM~\cite{OSLSM}, many few-shot segmentation methods have been proposed in recent years ~\cite{Co-FCN,AMP-2,PMM,FWB,PANet,CANet,PFENet,DAN,RePRI,SAGNN,FSOT,Zhang2020SGOneSG,CWT,ASGNet,HSNet}. Among others, an important challenge in few-shot segmentation is how to use support images towards capturing more meaningful information. Many recent state-of-the-art methods~\cite{Zhang2020SGOneSG,CANet,PFENet,PANet,yang2021mining,HSNet, VAT, ASNet} rely on feature masking~(FM)~\cite{Zhang2020SGOneSG} to discard irrelevant feature activations for reliable segmentation mask prediction. However, when masking a feature map, some crucial information in support images, such as the target object boundary, is partially lost. In particular, when the size of the target object is relatively small, this lost fine-grained spatial information renders it rather difficult to obtain accurate segmentation~(see Fig.~\ref{fig:teaser1}).

In this paper, we propose a simple, effective, and efficient technique to enhance feature masking~(FM)~\cite{Zhang2020SGOneSG}. We dub the enhanced FM as hybrid masking~(HM). In particular, we compensate for the loss of target object details in FM technique through leveraging a simple input masking~(IM) technique~\cite{OSLSM}. We note that IM is capable of preserving the fine details, especially around object boundaries, however, it lacks discriminative information, as such, after the removal of background information. To this end, we investigate the possibility of transferring object details in the IM to enrich FM technique. We instantiate the proposed hybrid masking~(HM) into two recent strong baselines: HSNet~\cite{HSNet} and VAT~\cite{VAT}. Results reveal more accurate segmentation masks by recovering the fine-grained details, such as target boundaries and target textures~(see Fig.~\ref{fig:teaser1}).
Following are the main contributions of this paper:
\begin{itemize}
    \item We propose a simple, effective,  and efficient way to enhance a de-facto feature masking technique~(FM) in several recent few-shot segmentation methods. 
    \item We perform extensive experiments to validate the effectiveness of proposed hybrid masking across two strong FSS baselines, namely HSNet~\cite{HSNet} and VAT~\cite{VAT} on three publicly available datasets: Pascal-5$^i$~\cite{OSLSM}, COCO-20$^i$~\cite{lin2015microsoft}, and FSS-1000~\cite{FSS1000}. 
    Results show notable improvements against the state-of-the-art methods in all datasets. %In particular, HMFS delivers an absolute gain of 3.6\% and 5.9\% over~\cite{zhang2021fewshot} with ResNet-50 and ResNet-101~\cite{Resnet} backbones, respectively under 1-shot setting on the challenging COCO-20$^i$~\cite{lin2015microsoft} dataset.
    \item We note that our HM facilitates improving the training efficiency. When integrated into HSNet~\cite{HSNet} with ResNet101~\cite{Resnet}, it speeds up its training convergence by around 11x times on average on COCO-20$^i$~\cite{lin2015microsoft}.
    
\end{itemize}

\section{Related Work}

\noindent \textbf{Few-shot segmentation.} 
The work of Shaban \textit{et al.}~\cite{OSLSM} is believed to introduce the few shot segmentation task to the community. It generated segmentation parameters by using the conditioning branch on the support set. 
Later, we observe steady progress in this task, and so several methods were proposed \cite{Co-FCN,AMP-2,PMM,FWB,PANet,CANet,PFENet,DAN,RePRI,SAGNN,FSOT,Zhang2020SGOneSG,CWT,ASGNet,HSNet}. CANet~\cite{CANet} modified the cosine similarity with the additive alignment module and enhanced the performance by performing various iterations. To improve segmentation quality, PFENet~\cite{PFENet} designed a pyramid module and used a prior map. Inspired by prototypical networks \cite{snell2017prototypical}, PANet~\cite{PANet} leveraged novel prototype alignment network. Along similar lines, PPNet~\cite{PPNet} utilized part-aware prototypes to get the detailed object features and PMM~\cite{PMM} used the expectation-maximization~(EM) algorithm to generate multiple prototypes. ASGNet~\cite{ASGNet} proposed two modules, superpixel-guided clustering~(SGC) and guided prototype allocation~(GPC) to extract and allocate multiple prototypes. In pursuit of improving correspondence between support and query images,  DAN~\cite{DAN} democratized graph attention. Yang \textit{et al.}~\cite{yang2021mining} introduced a method to mine latent classes in the background, and CWT~\cite{CWT} designed a simple classifier with transformer. CyCTR~\cite{zhang2021fewshot} mined information from the whole support image using transformer. ASNet~\cite{ASNet} proposed the integrative few-shot learning framework~(iFSL) overcoming limitations of few-shot classification and few-shot segmentation. HSNet~\cite{HSNet} utilized efficient 4D convolution to analyze deeply accumulated features and achieved remarkable performance. Recently, VAT~\cite{VAT} proposed a cost aggregation network, based on transformers, to model dense semantic correspondence between images and capture intra-class variations. We validate the effectiveness of our hybrid masking approach by instantiating it in two strong FSS baselines: HSNet~\cite{HSNet} and VAT~\cite{VAT}.

\noindent \textbf{Feature masking.}
Zhang~\textit{et al.}~\cite{Zhang2020SGOneSG} proposed Masked Average Pooling~(MAP) to eliminate irrelevant feature activations which facilitates reliable mask prediction. In MAP, feature masking~(FM) was introduced and utilized before average pooling. Afterward, FM was widely adopted as the de-facto technique to achieve feature masking~\cite{Zhang2020SGOneSG,CANet,PFENet,PANet,yang2021mining,HSNet, VAT, ASNet}. We note that the FM method loses information about the target object in the process of feature masking. Specifically, it is prone to losing the fine-grained spatial details, which can be crucial for generating a precise segmentation mask. In this work, we compensate for the loss of target object details in FM technique via leveraging a simple input masking~(IM) technique~\cite{OSLSM}.

\noindent \textbf{Input masking.} Input masking~(IM)~\cite{OSLSM} is a technique to eliminate background pixels by multiplying the support image with its corresponding support mask. There were two key motivations behind erasing the background pixels. First, the largest object in the image has a tendency to dominate the network. Second, the variance of the output parameters increased when the background information was included in the input. We observe that IM can preserve the fine details, however, it lacks target discriminative information, important for distinguishing between the foreground and the background. In this work, we investigate the possibility of transferring object details present in the IM to enrich the FM technique, thereby exploiting the complementary strengths of both.

\section{Methodology}
\begin{figure}[t]
\centering
\includegraphics[height=6.0cm]{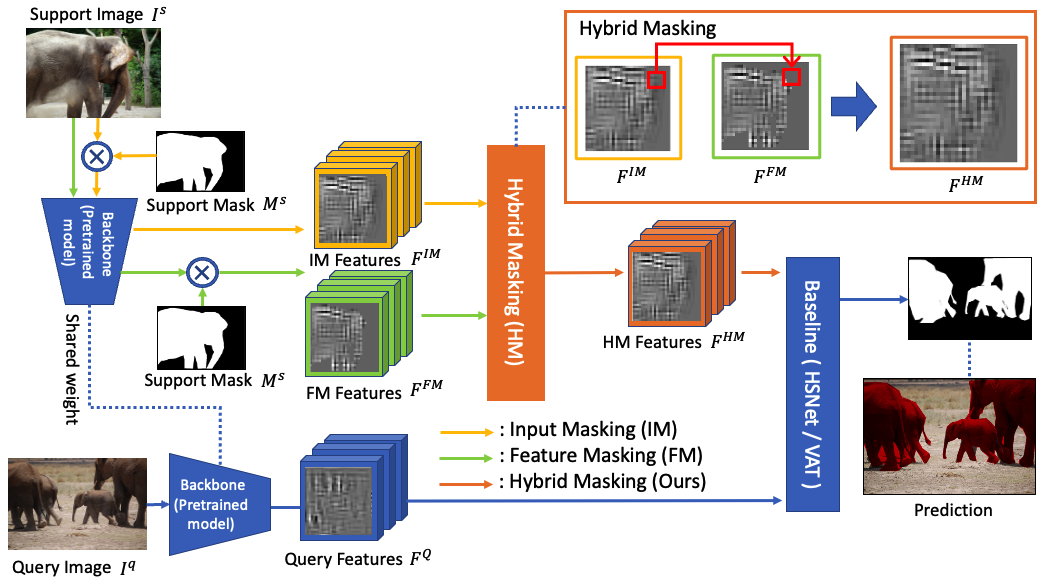}
\caption{ \small \textbf{The overall architecture when our proposed hybrid masking (HM) approach is integrated into FSS baselines.} At its core, it contains a feature backbone, a feature masking~(FM) technique. After extracting support and query features, the feature masking suppresses irrelevant activations in the support features. We introduce a simple, effective, and efficient way to enhance feature masking~(FM), termed as \emph{hybrid masking}~(HM). It compensates for the loss of target object details in the FM technique by leveraging a simple input masking~(IM) technique~\cite{OSLSM}.}

\label{fig:overall_architecture}
\end{figure}

Fig.~\ref{fig:overall_architecture} displays the overall architecture when our proposed \emph{hybrid masking} (HM) approach is introduced into the FSS baselines, such as HSNet~\cite{HSNet} and VAT~\cite{VAT}. Fundamentally, it comprises of a feature backbone for extracting support and query features, a feature masking~(FM) technique for suppressing irrelevant support activations, and FSS model (i.e. HSNet/VAT) for predicting the segmentation mask from the relevant activations. In this work, we propose a simple, effective, and efficient way to enhance feature masking~(FM), termed as hybrid masking~(HM). It compensates for the loss of target object details in the FM technique by leveraging a simple input masking~(IM) technique~\cite{OSLSM}. 
In what follows, we first lay out the problem setting~(sec.~\ref{subsection:problem_setting}), next we describe feature masking technique~(sec.~\ref{subsection:feature_masking}), and finally we detail the proposed hybrid masking for few-shot segmentation~(sec.~\ref{subsection:hmfs}).

\subsection{Problem Setting}
\label{subsection:problem_setting}
% The goal of few-shot segmentation is to train a model capable of segmenting the target object in a query image given a few annotated images from the target class.
Few-shot segmentation's objective is to train a model that can recognize the target object in a query image given a small sample of annotated images from the target class.
We tackle this problem using the widely adopted episodic training scheme~\cite{episodic, PANet, HSNet, VAT}, which has been shown to reduce overfitting. We have the disjoint sets of training classes $C_{train}$ and testing classes $C_{test}$. %
The training data $D_{train}$ belongs to $C_{train}$ and the testing data $D_{test}$ is from $C_{test}$.
Multiple episodes are constructed using the $D_{train}$ and $D_{test}$. 
A support set, $S = (I^s, M^s)$, and a query set, $Q = (I^q, M^q)$, are the two components that make up each episode. $I$ and $M$ represent an image and its mask. We have $N_{train}$ episodes for training $D_{train} = \{(S_i, Q_i)\}^{N_{train}}_{i=1} $ and $N_{test}$ episodes for testing $D_{test} = \{(S_i, Q_i)\}^{N_{test}}_{i=1}$. 
% where Ntrain and Ntest represent the training and testing episode counts.
% where $N_{train}$ is the number of episodes for training and $N_{test}$ denotes the number of episodes for testing.
%
% Sampled episodes from $D_{train}$ are used to train a model to predict query mask $M^q$.
Sampled episodes from $D_{train}$ are used to train a model to predict query mask $M^q$.
Afterward, the learned model is evaluated by randomly sampling episodes from the testing data $D_{test}$ in the same manner and comparing the predicted query masks to the ground truth.

\begin{wrapfigure}[20]{h}[0cm]{0.42\textwidth}\centering
\vspace{-30pt}    
%    \vspace{-\normalbaselineskip}
    \includegraphics[width=0.42\textwidth,height=11\normalbaselineskip]{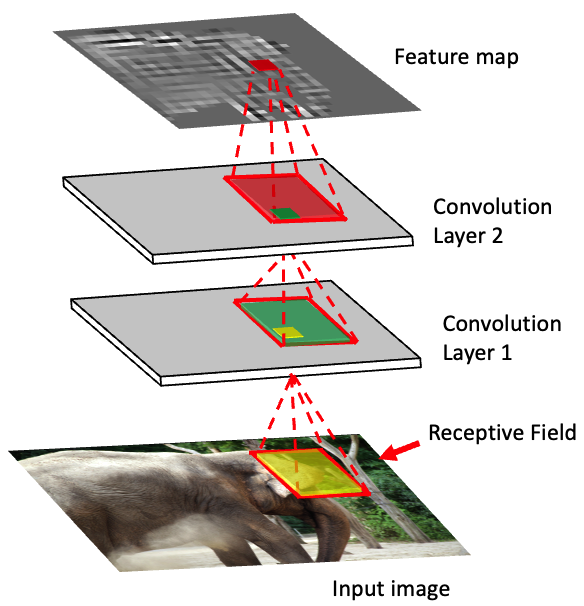}
% \vspace{3pt}    
\caption{ \small Impact of growing receptive field on feature masking. The input image's elephant is a target object and the other pixels are background. One pixel at the feature is generated from lots of pixels' information from previous layer. The background and target object information both can be present in one feature map pixel.}    
\label{fig:receptive_field}
\end{wrapfigure}

\subsection{Feature Masking}
\label{subsection:feature_masking}
Zhang \textit{et al.}~\cite{Zhang2020SGOneSG} argued that IM greatly increases the variance of the input data for a unified network, and according to Long \textit{et al.}~\cite{FCN}, the relative positions of input pixels can be preserved by fully convolutional networks.
These two ideas motivated Masked Average Pooling~(MAP), which ideally extracts the features of a target object while eliminating the background content.
Although MAP falls short of this in practice, it remains helpful for learning better object features~\cite{Deeplab} while keeping the input structure of the network unchanged.  In particular, the feature masking part is still widely used.

Given a support RGB image $I^s \in \mathbb{R}^{3 \times w \times h }$ and a support mask $\mathbf{M}^s \in \{0,1\}^{w \times h}$, where w and h are the width and height of the image, the support feature maps of $I^s$ are  $F^s \in \mathbb{R}^{c \times w' \times h' }$, where c is the number of channels, $w'$ and $h'$ are the width and height of the feature maps. Feature masking is performed after matching the mask to the feature size using the bilinear interpolation. We denote a function resizing the mask as $\tau (\cdot) : \mathbb{R}^{ w \times h }$ $ \rightarrow \mathbb{R}^{c \times w' \times h' }  $ . Then, the feature masking features $F^{FM}$$\in \mathbb{R}^{c \times w' \times h'}$ are computed according to Eqn.~\ref{equ:1},
\begin{align} \label{equ:1}
    F^{FM} = F^s \odot \tau (M^{s}) ,
\end{align}
where $\odot$ denotes the Hadamard product. Zhang \textit{et al.}~\cite{Zhang2020SGOneSG} fit the feature size to the mask size, but we conversely fit the mask to the feature size.

Feature masking (FM), which forms the core part of Masked Average Pooling~(MAP)~\cite{Zhang2020SGOneSG}, is utilized to eliminate background information from support features and has become the de facto technique for masking feature maps, appearing in several recent few-shot segmentation methods \cite{Zhang2020SGOneSG,CANet,PFENet,PANet,yang2021mining} even in current state-of-the-art~\cite{HSNet}.
However, FM inadvertently eliminates both the background and the target object information because one pixel from the last layer's feature map corresponds to many pixels in the input image.

Fig.~\ref{fig:receptive_field} shows that one pixel of the feature map could contain background and target object information together~\cite{receptive_analysis}. This is further analyzed in Fig.~\ref{fig:feature_difference}, which clearly shows that FM loses useful information through its masking and progressively worsens with deeper layers.
If a target object in the support set appears very small, the segmentation of the query image becomes even more challenging because the features are fed into network with relatively large proportion of undesired background features. Fig.~\ref{fig:comparison} shows the limitation of FM.

\subsection{Hybrid Masking for Few-shot Segmentation~(HM)}
\label{subsection:hmfs}

We aim to maximize target information from the support set so that the network can efficiently learn to provide more accurate segmentation of a target object.
%more accurately in an  learn to segment a target object.

The overall architecture when our proposed HM is integrated into a FSS baseline is shown in Fig.~\ref{fig:overall_architecture}. We obtain two feature maps, $F^{IM}$ and $F^{FM}$, using IM and FM respectively. These two feature maps are merged by hybrid masking~(Alg.~\ref{alg:two}) to generate HM feature map $F^{HM}$. This HM feature map will be used as input for HSNet and VAT, which takes full advantage of the given features to predict the target object mask.

% \st{Following the work of FPN}~\cite{FPN,HSNet}, \st{we extract these features maps using deep convolutional layers and stack them to create 4D correlation tensors. We propose hybrid masking to merge these feature maps and find the correlation between the support features and query features using cosine similarity. This correlation set is fed into the encoder and the decoder, which takes full advantage of the given features to predict the target object mask.}

\noindent \textbf{Input Masking~(IM).} IM~\cite{OSLSM} eliminates background pixels by multiplying the support image with its corresponding support mask because of two empirical reasons.~(1)~The network has a tendency to favor the largest object in the image, which is not the object we want to segment.~(2)~The background information will result in an increase in the variance of the output parameters.

Suppose we have the RGB support image $I^s \in \mathbb{R}^{3 \times w \times h}$
and a support mask $\mathbf{M}^s \in \{0,1\}^{w \times h}$ in the image space, where $w$ and $h$ are the width and height of the image. IM, computed as
\begin{align}  \label{eq:2}
  I^{s'} & =  I^s \odot \tau (M^s)  , 
\end{align}
contains the target object alone. We use the function $\tau ( \cdot ) $ for resizing the mask $M^s$ to fit the image $I^s$.

\noindent \textbf{Hybrid Masking~(HM).}
We propose an alternative masking approach, which takes advantage of the features generated by both FM and IM.
First, FM and IM features are computed according to the existing methods. The unactivated values in the FM features are then replaced with IM features. Other activated values remain without replacing to maintain FM features. We name this process as hybrid masking~(Alg.~\ref{alg:two}).
HM prioritizes the information from FM features and supplements the lacking information, such as the precise target boundaries and fine-grained texture information, from IM features, which is superior for delineating the boundaries of target objects and the missing texture information. The method is as follows even if feature maps are stacked to have a sequence of intermediate feature maps. Only one more loop is needed for the deep feature maps.

\vspace{-6mm}
\begin{algorithm}
\SetKwInOut{Input}{Input}
\SetKwInOut{Output}{Output}

\caption{Hybrid Masking}\label{alg:two}
\Input{IM feature maps $F^{IM}$ and FM features maps $F^{FM}$}

Each channel $i$, $f^{IM}_i$ $\in$ $F^{IM}$ and $f^{FM}_i$ $\in$ $F^{FM}$

\For{i =1, \ldots, c}{

Set  $f^{HM}_i = f^{FM}_i$

\For{Entire pixels $\in$ $f^{HM}_i$  }{
Find an inactive pixel, $p$ $\in$ $f^{HM}_i$ 

\If{$p \leq 0$}{
%\hl{VP: what does "pixel not larger than zero" mean? I would also refrain from using "not larger" and instead say "pixel smaller than zero".}
Replace the pixel, $p$, with corresponding pixel $\in$ $f^{IM}_i$
     }

}

}
\Output{HM feature maps $F^{HM}$}
\end{algorithm}
\vspace{-6mm}

% \subsubsection{Cosine similarity.}
% The cosine similarity is computed by comparing all pixel values between HM features and query features.  Therefore, even if the position of the target object in the query image differs from the positions of target objects in the support set, the cosine similarity is unaffected. Suppose we have HM features $F^{HM} \in \mathbb{R}^{c \times w' \times h' } $ and query features $F^{Q} \in \mathbb{R}^{c \times w' \times h' } $.
% We resize and transpose these two features into $F^{HM'} \in \mathbb{R}^{c \times (w' \cdot h') } $ and $F^{Q'} \in \mathbb{R}^{ (w' \cdot h') \times c }$, and construct the correlation set $C  \in \mathbb{R}^{ (w' \cdot h') \times (w' \cdot h') }$ %using Eqn.~\ref{eq:3}.
% as
% \begin{align} \label{eq:3}
%   C = ReLU \bigg(\frac{F^{Q'} \cdot F^{HM'}}{ \Vert F^{Q'} \Vert \Vert F^{HM'} \Vert  } \bigg).
% \end{align}
% Then, we resize again to get a 4D correlation set $C'  \in \mathbb{R}^{ w' \times  h' \times w' \times h' }$.

%\subsubsection{Encoder and decoder.}

The generated HM feature maps are used as inputs to two strong FSS models~(HSNet and VAT). These two models are best-suited for fully utilizing the features generated by hybrid masking, because they build multi-level correlation maps taking advantage of the rich semantics that are provided at different feature levels.

% is able to analyze very deep features with efficient 4D convolutions. The 4D correlation set $C'$ is fed into this last component.

% exploit rich semantics present in different feature levels, and build multi-level correlation maps.

\section{Experiment}
\subsection{Setup}
\subsubsection{Datasets.}
We evaluate the efficacy of our hybrid masking technique on three publicly available segmentation benchmarks: PASCAL-5$^i$~\cite{OSLSM} , COCO-20$^i$~\cite{lin2015microsoft}, and FSS-1000~\cite{FSS1000}. PASCAL-5$^i$ was produced from PASCAL VOC 2012~\cite{pascal} with additional mask annotations~\cite{pascal_2}. PASCAL-5$^i$ contains 20 types of object classes, COCO-20$^i$ contains 80 classes, and FSS-1000 contains 1000 classes. The PASCAL and COCO data sets were divided into four folds following the training and evaluation methods of other works~\cite{PPNet,FWB,PFENet,DAN,PMM, HSNet, VAT, ASNet}, where each fold of PASCAL-5$^i$ consisted of 5 classes, and each fold of COCO-20$^i$ had 20. We conduct cross-validation using these four folds. When evaluating a model on fold$^i$, all other classes not belonging to fold$^i$ are used for training. 1000 episodes are sampled from the other  fold$^i$ to evaluate the trained model. For FSS-1000, the training, validation, and test datasets are divided into 520, 240, and 240 classes.

\subsubsection{Implementation details.} We integrate our hybrid masking technique into two FSS baselines: HSNet~\cite{HSNet} and VAT~\cite{VAT}, and the resulting methods are denoted by HSNet-HM and VAT-HM.
We use ResNet50~\cite{Resnet} and ResNet101~\cite{Resnet} backbone networks pre-trained on ImageNet~\cite{image_net} with their weights frozen to extract features, following HSNet~\cite{HSNet} and VAT~\cite{VAT}. From conv3$\_$x to conv5$\_$x of ResNet~(i.e., the three layers before global average pooling), the features in the bottleneck part before the ReLU activation of each layer were stacked up to create deep features. We follow the HSNet~\cite{HSNet} and the VAT~\cite{VAT} default settings for optimizer~\cite{Adam} and learning rate. A batch size of 20 is used for HSNet-HM training for all benchmarks. For VAT-HM training, 8, 4, and 4 batch sizes are utilized for COCO-20$^i$, PASCAL-5$^i$ and FSS-1000 respectively. 
%\hl{VP: have you tried running experiments on seqamgpu02, with A6000?  48GB vs 16GB memory.} \textcolor{red}{ah yes but I was in hurry so I used small size GPU as well so I need to use the same batch size like 16gb. I will mention about that}
%This reduced batch size shows negligible impact on the performance.
We used data augmentation for HSNet-HM training on PASCAL-5$^i$ following \cite{data_aug_1, data_gug_2, VAT}. For COCO-20$^i$ and FSS-1000 benchmarks, no data augmentation was employed when training HSNet-HM. 

\subsubsection{Evaluation metrics.}
Following \cite{HSNet,PFENet,DAN,VAT}, we adopt two evaluation metrics, mean intersection over union~(mIoU) and foreground-background IoU~(FB-IoU) for model evaluation. The mIoU averages the IoU values for all classes in each fold. FB-IoU calculates the foreground and background IoU values ignoring object classes and averages them. Note that, mIoU is a better indicator of model generalization than FB-IoU~\cite{HSNet}.

\subsection{Comparison with the State-of-the-Art~(SOTA)}

\subsubsection{PASCAL-5$^i$.} Table \ref{table:performance_pascal} compares our methods, HSNet-HM and VAT-HM, with other methods on PASCAL-5$^i$ datasets. In the 1-shot test, HSNet-HM provides a gain of 0.7$\%$ mIoU compared to HSNet~\cite{HSNet} with ResNet50 backbone and performs on par with HSNet~\cite{HSNet} using ReNet101 backbone. In the 5-shot test, HSNet-HM shows slightly inferior performance in mIoU and FB-IoU. VAT-HM shows a similar pattern to HSNet-HM. In the 1-shot test, VAT-HM shows a gain of 0.5$\%$ mIoU with ResNet50 and a gain of 0.3$\%$ mIoU with ResNet101. In the 5-shot test, VAT-HM provides an improvement of 0.8$\%$ mIoU with ResNet50.% but no improvement with ResNet101.

% Attention(ours)  69.3 72.7  61.3 65.2  67.1.     

\subsubsection{COCO-20$^i$.}
Table \ref{table:performance_coco} reports results on the COCO-20$^i$ dataset. In 1-shot test, HSNet-HM, VAT-HM, and ASNet-HM show a significant improvement over HSNet~\cite{HSNet}, VAT~\cite{VAT}, and ASNet~\cite{ASNet}. HSNet-HM delivers a gain of 5.1$\%$ and 5.3$\%$ in mIoU with ResNet50 and ResNet101 backbones, respectively. VAT-HM provides a gain of 2.9$\%$ mIoU with ResNet50. ASNet-HM provides a gain of 2.5$\%$ and 2.8$\%$ mIoU with ResNet50 and ResNet101. Similarly, in 5-shot test, HSNet-HM outperforms HSNet~\cite{HSNet} by 3.5$\%$ and 1.1$\%$ with ResNet50 and ResNet101 backbones, respectively. VAT-HM provides 0.4$\%$ mIoU improvement with ResNet50 in 5-shot test. ASNet-HM shows slightly worse performance with ResNet50 but ASNet-HM delivers a gain of 1.1$\%$ with ResNet101. Fig.~\ref{fig:teaser1} draws visual comparison with HSNet~\cite{HSNet} and VAT~\cite{VAT} under several challenging segmentation instances. Note that, compared to HSNet and VAT, HSNet-HM and VAT-HM produce more accurate segmentation masks that recover fine-grained details under appearance variations and complex backgrounds.

\begin{table}[!htbp]
\caption{Performance comparison with the existing methods on Pascal-5$^i$~\cite{pascal}. Superscript asterisk denotes that data augmentation was applied during training. Best results are bold-faced and the second best are underlined.
%\hl{VP: VAT-HM 78.5 on ResNet50 is not best performance.}
} %An explanation of the performance degradation can be found in Section 5.}
\label{table:performance_pascal}

\centering
\resizebox{\textwidth}{!}{%
\begin{tabular}{@{}cc|cccccc|cccccc@{}}
\toprule
\multirow{2}{*}{\begin{tabular}[c]{@{}c@{}}Backbone\\ feature\end{tabular}} & \multirow{2}{*}{Methods} & \multicolumn{6}{c|}{1-shot} & \multicolumn{6}{c}{5-shot}  
\\
 &  & $5^0$ & $5^1$ & $5^2$ & $5^3$ & mIoU & FB-IoU & $5^0$ & $5^1$ & $5^2$ & $5^3$ & mIoU & FB-IoU  \\ \midrule
\multirow{11}{*}{ResNet50~\cite{Resnet}} & PANet~\cite{PANet} & 44.0 & 57.5 & 50.8 & 44.0 & 49.1 & - & 55.3 & 67.2 & 61.3 & 53.2 & 59.3 & - \\
 & PFENet~\cite{PFENet} & 61.7 & 69.5 & 55.4 & 56.3 & 60.8 & 73.3 & 63.1 & 70.7 & 55.8 & 57.9 & 61.9 & 73.9 \\
 & ASGNet~\cite{ASGNet} & 58.8 & 67.9 & 56.8 & 53.7 & 59.3 & 69.2 & 63.4 & 70.6 & 64.2 & 57.4 & 63.9 & 74.2 \\
 & CWT~\cite{CWT} & 56.3 & 62.0 & 59.9 & 47.2 & 56.4 & - & 61.3 & {68.5} & 68.5 & 56.6 & 63.7 & - \\
 & RePRI~\cite{RePRI} & 59.8 & 68.3 & {62.1} & 48.5 & 59.7 & - & 64.6 & 71.4 & \textbf{71.1} & 59.3 & 66.6 & - \\
 & CyCTR~\cite{zhang2021fewshot} & {67.8} & \textbf{72.8} & 58.0 & 58.0 & 64.2 & - & {71.1} & \underline{73.2} & 60.5 & 57.5 & 65.6 & - \\ 
 & HSNet~\cite{HSNet} & 64.3 & 70.7 & 60.3 & 60.5 & 64.0 & {76.7} & 70.3 & \underline{73.2} & 67.4 & \textbf{67.1} & {69.5} & \underline{80.6} \\
& HSNet$^\ast$ & 63.5 & 70.9 & \underline{61.2} & {60.6} & {64.3} & \textbf{78.2} & {70.9} & {73.1} & 68.4 & {65.9} & \underline{69.6} & \underline{80.6} \\ 
& VAT~\cite{VAT} & 67.6 & \underline{71.2} & \textbf{62.3} & 60.1 & \underline{65.3} & \underline{77.4} & \underline{72.4} & \textbf{73.6} & \underline{68.6} & {65.7} & \textbf{70.0} & \textbf{80.9} \\
 \cmidrule(l){2-14} 
% 72.9, 73.2,, 67,9, 68.4

  & HSNet$^\ast$-HM & \textbf{69.0} & {70.9} & 59.3 & \underline{61.0} & {65.0} & 76.5  & 69.9 & 72.0 & 63.4 & 63.3 & 67.1  & {77.7}  \\ 
  & VAT-HM & \underline{68.9} & {70.7} & {61.0} & \textbf{62.5} & \textbf{65.8}  & {77.1} & \textbf{71.1} & \underline{72.5} & 62.6 & \underline{66.5} & {68.2} & {78.5} \\
  \midrule  
  
\multirow{12}{*}{ResNet101~\cite{Resnet}} & FWB~\cite{FWB} & 51.3 & 64.5 & 56.7 & 52.2 & 56.2 & - & 54.8 & 67.4 & 62.2 & 55.3 & 59.9 & - \\
 & DAN~\cite{DAN} & 54.7 & 68.6 & 57.8 & 51.6 & 58.2 & 71.9 & 57.9 & 69.0 & 60.1 & 54.9 & 60.5 & 72.3 \\
 & PFENet~\cite{PFENet} & 60.5 & 69.4 & 54.4 & 55.9 & 60.1 & 72.9 & 62.8 & 70.4 & 54.9 & 57.6 & 61.4 & 73.5 \\
 & ASGNet~\cite{ASGNet} & 59.8 & 67.4 & 55.6 & 54.4 & 59.3 & 71.7 & 64.6 & 71.3 & 64.2 & 57.3 & 64.4 & 75.2  \\
 & CWT~\cite{CWT} & 56.9 & 65.2 & 61.2 & 48.8 & 58.0 & - & 62.6 & 70.2 & \textbf{68.8} & 57.2 & 64.7 & - \\
 & RePRI~\cite{RePRI} & 59.6 & 68.6 & {62.2} & 47.2 & 59.4 & - & 66.2 & 71.4 & 67.0 & 57.7 & 65.6 & -  \\
 & CyCTR~\cite{zhang2021fewshot} & {69.3} & \textbf{72.7} & 56.5 & 58.6 & 64.3 & 72.9 & \underline{73.5} & 74.0 & 58.6 & 60.2 & 66.6 & 75.0 \\
 & HSNet~\cite{HSNet} & 67.3 & {72.3} & 62.0 & 63.1 & {66.2} & 77.6 & 71.8 & {74.4} & 67.0 & {68.3} & {70.4} & {80.6} \\
 & HSNet$^\ast$ & 67.5 & \textbf{72.7} & \underline{63.5} & {63.2} & {66.7} & {77.7} & 71.7 & {74.8} & {68.2} & \underline{68.7} & {70.8} & {80.9} \\
& VAT~\cite{VAT} & {68.4} & \underline{72.5} & \textbf{64.8} & {64.2} & \underline{67.5}  & \underline{78.8} & {73.3} & \underline{75.2} & \underline{68.4} & \textbf{69.5} & \textbf{71.6} & \textbf{82.0} \\

 \cmidrule(l){2-14}  
 & HSNet$^\ast$-HM & \underline{69.8} & 72.1 & 60.4 & \underline{64.3} & {66.7}  & {77.8} & {72.2} & 73.3 & 64.0 & 67.9 & 69.3 & 79.7 \\ 
 & VAT-HM & \textbf{71.2} & \textbf{72.7} & {62.7} & \textbf{64.5} & \textbf{67.8}  & \textbf{79.4} & \textbf{74.0} & \textbf{75.5} & 65.4 & {68.6} & \underline{70.9} & \underline{81.5} \\

 \bottomrule

\end{tabular}
}
\end{table}

\begin{table}
\caption{Performance comparison on COCO-20$^i$~\cite{lin2015microsoft} in mIoU and FB-IoU. Best results are bold-faced and the second best are underlined.
}
\label{table:performance_coco}
\centering
\scalebox{0.9}{%
\begin{tabular}{@{}cc|cccccc|cccccc@{}}
\toprule
\multirow{2}{*}{\begin{tabular}[c]{@{}c@{}}Backbone\\ feature\end{tabular}} & \multirow{2}{*}{Methods} & \multicolumn{6}{c|}{1-shot} & \multicolumn{6}{c}{5-shot} 
\\
 &  & $20^0$ & $20^1$ & $20^2$ & $20^3$ & mIoU & FB-IoU & $20^0$ & $20^1$ & $20^2$ & $20^3$ & mIoU & FB-IoU \\ \midrule
\multirow{12}{*}{ResNet50~\cite{Resnet}} & PMM~\cite{PMM} & 29.3 & 34.8 & 27.1 & 27.3 & 29.6 & - & 33.0 & 40.6 & 30.3 & 33.3 & 34.3 & - \\
 & RPMM~\cite{PMM} & 29.5 & 36.8 & 28.9 & 27.0 & 30.6 & - & 33.8 & 42.0 & 33.0 & 33.3 & 35.5 & -\\
 & PFENet~\cite{PFENet} & 36.5 & 38.6 & 34.5 & 33.8 & 35.8 & - & 36.5 & 43.3 & 37.8 & 38.4 & 39.0 & - \\
 & ASGNet~\cite{ASGNet} & - & - & - & - & 34.6 & 60.4 & - & - & - & - & 42.5 & 67.0\\
 & RePRI~\cite{RePRI} & 32.0 & 38.7 & 32.7 & 33.1 & 34.1 & - & 39.3 & 45.4 & 39.7 & 41.8 & 41.6 & - \\
 & HSNet~\cite{HSNet} & 36.3 & {43.1} & 38.7 & 38.7 & 39.2 & {68.2} & {43.3} & {51.3} & {48.2} & {45.0} & {46.9} & {70.7} \\
 & CyCTR~\cite{zhang2021fewshot} & {38.9} & 43.0 & {39.6} & {39.8} & {40.3} & - & 41.1 & 48.9 & 45.2 & {47.0} & 45.6 & - \\
   & VAT~\cite{VAT} & {39.0} & 43.8 & 42.6 & {39.7} & {41.3}  & {68.8} & {44.1} & 51.1 & 50.2 & 46.1 & 47.9 & \underline{72.4} \\
& ASNet~\cite{ASNet} & {41.5} & 44.1 & 42.8 & {40.6} & {42.2}  & {69.4} & \textbf{48.0} & \underline{52.1} & 49.7 & \underline{48.2} & \textbf{49.5} & \textbf{72.7} \\ 
\cmidrule(l){2-14} 
 & HSNet-HM&  {41.0}&  \underline{45.7}&  \textbf{46.9}& \underline{43.7} & \underline{44.3} & \textbf{70.8} & {45.3} & \textbf{53.1}  & \textbf{52.1} & {47.0}  &  \underline{49.4} &  72.2  \\
& VAT-HM & \underline{42.2} & {43.3} & \underline{45.0} & {42.2} & {43.2}  & {70.0} & {45.2} & {51.0} & \underline{50.7} & {46.4} & 48.3 & 71.8 \\
& ASNet-HM&  \textbf{42.8}&  \textbf{46.0}&  {44.8}& \textbf{45.0} & \textbf{44.7} & \underline{70.4} & \underline{46.3} & {50.2}  & {48.4} & \textbf{48.6}  &  {48.4} &  72.2  \\
\midrule
 
\multirow{7}{*}{ResNet101~\cite{Resnet}} & FWB~\cite{FWB} & 17.0 & 18.0 & 21.0 & 28.9 & 21.2 & - & 19.1 & 21.5 & 23.9 & 30.1 & 23.7 & - \\
 & DAN~\cite{DAN} & - & - & - & - & 24.4 & 62.3 & - & - & - & - & 29.6 & 63.9  \\
 & PFENet~\cite{PFENet} & {36.8} & 41.8 & {38.7} & 36.7 & 38.5 & 63.0 & 40.4 & 46.8 & 43.2 & 40.5 & 42.7 & 65.8 \\
 & HSNet~\cite{HSNet} & 37.2 & {44.1} & {42.4} & {41.3} & {41.2} & {69.1} & {45.9} & \underline{53.0} & \underline{51.8} & {47.1} & \underline{49.5} & {72.4} \\
& ASNet~\cite{ASNet} & \underline{41.8} & 45.4 & 43.2 & {41.9} & {43.1}  & {69.4} & \textbf{48.0} & 52.1 & 49.7 & 48.2 & \underline{49.5} & 72.7 \\
\cmidrule(l){2-14} 
 & HSNet-HM & {41.2} & \textbf{50.0} & \textbf{48.8} & \underline{45.9} & \textbf{46.5} & \textbf{71.5}  & {46.5}  & \textbf{55.2} & \underline{51.8}  & \underline{48.9}  & \textbf{50.6} & \underline{72.9} \\
   & ASNet-HM & \textbf{43.5} & \underline{46.4} & \underline{47.2} & \textbf{46.4} & \underline{45.9} & \underline{71.1}  & \underline{47.7}  & {51.6} & \textbf{52.1}  & \textbf{50.8}  & \textbf{50.6} & \textbf{73.3} \\
  \bottomrule
\end{tabular}
}

\end{table}

\begin{table}[!htp]
\caption{Performance comparison with other methods on FSS-1000~\cite{FSS1000} dataset. Best results are bold-faced and the second best are underlined.}
\label{table:performance_fss}

\centering
\scalebox{0.8}{

\begin{tabular}{@{}llll|llll@{}}
\toprule
\multirow{2}{*}{\begin{tabular}[c]{@{}l@{}}Backbone\\ feature\end{tabular}} & \multirow{2}{*}{Methods} & \multicolumn{2}{l|}{mIoU} & \multirow{2}{*}{\begin{tabular}[c]{@{}l@{}}Backbone\\ feature\end{tabular}} & \multirow{2}{*}{Methods} & \multicolumn{2}{l}{mIoU} \\
 &  & 1-shot & 5-shot &  &  & 1-shot & 5-shot \\ \midrule
\multirow{5}{*}{ResNet50~\cite{Resnet}} & FSOT~\cite{FSOT} & 82.5 & 83.8 & \multirow{5}{*}{ResNet101~\cite{Resnet}} & DAN~\cite{DAN} & 85.2 & 88.1 \\
 & HSNet~\cite{HSNet} & {85.5} & 87.8 &  & HSNet~\cite{HSNet} & 86.5 & 88.5 \\ 
  & VAT~\cite{VAT} & {\textbf{89.5}} & \textbf{90.3} & & VAT~\cite{VAT} & \underline{90.0}  & \textbf{90.6} \\
 \cmidrule(lr){2-4}
 \cmidrule(l){6-8} 
 & HSNet-HM & 87.1 & 88.0 &  & HSNet-HM & 87.8 & 88.5 \\ 
    & VAT-HM & \underline{89.4} & \underline{89.9} & &VAT-HM& \textbf{90.2} & \underline{90.5} \\ 

 \bottomrule
\end{tabular}
}

\end{table}

\begin{table}[!htp]
\caption{Comparison of generalization performance with domain shift test.  A model was trained on COCO-20$^i$~\cite{lin2015microsoft} and then evaluated on PASCAL-5$^i$~\cite{pascal}.
%\hl{VP: no results with VAT here?} \textcolor{red}{Because VAT paper didn't mention this experiment in their paper. But HSNet showed this experiment }
}
\label{table:performance_pascal_shift}
\centering
\scalebox{0.8}{%
\begin{tabular}{@{}cc|ccccc|ccccc@{}}
\toprule
\multirow{2}{*}{\begin{tabular}[c]{@{}c@{}}Backbone\\ feature\end{tabular}} & \multirow{2}{*}{Methods} & \multicolumn{5}{c|}{1-shot} & \multicolumn{5}{c}{5-shot}  \\
  &  & $5^0$ & $5^1$ & $5^2$ & $5^3$ & mIoU & $5^0$ & $5^1$ & $5^2$ & $5^3$ & mIoU    \\ \midrule
\multirow{7}{*}{ResNet50~\cite{Resnet}} & RPMM~\cite{PMM} & 36.3 & 55.0 & 52.5 & 54.6 & 49.6 & 40.2 & 58.0 & 55.2 & 61.8 & 53.8   \\
 & PFENet~\cite{PFENet} & 43.2 & \underline{65.1} & {66.5} & 69.7 & 61.1 & 45.1 & 66.8 & 68.5 & 73.1 & 63.4   \\
 & RePRI~\cite{RePRI} & \textbf{52.2} & 64.3 & 64.8 & 71.6 & {63.2}  & {56.5} & \underline{68.2} & 70.0 & 76.2 & 67.7   \\
 & HSNet~\cite{HSNet} & {45.4} & 61.2 & 63.4 & {75.9} & 61.6  & \underline{56.9} & 65.9 & {71.3} & {80.8} & \underline{68.7}  \\
& VAT& \underline{52.1} & 64.1 & 67.4 & 74.2 & 64.5 & \textbf{58.5} & 68.0 & \underline{72.5} & 79.9 & \textbf{69.7} \\  
 \cmidrule(l){2-12} 
  & HSNet-HM& 43.4 & \textbf{68.2} & \textbf{69.4} & \textbf{79.9} & \textbf{65.2} & 50.7 & \textbf{71.4} & \textbf{73.4} & \textbf{83.1} & \textbf{69.7} \\  
& VAT-HM& 48.3 & 64.9 & \underline{67.5} & \underline{79.8} & \underline{65.1} & 55.6 & 68.1 & 72.4 & \underline{82.8} & \textbf{69.7} \\  
   \midrule  
%   \midrule
  
\multirow{2}{*}{ResNet101~\cite{Resnet}}  & HSNet~\cite{HSNet} & \textbf{47.0} & \underline{65.2} & \underline{67.1} & \underline{77.1} & \underline{64.1} & \textbf{57.2} & \underline{69.5} & \underline{72.0} & \underline{82.4} & \underline{70.3}\\
 \cmidrule(l){2-12} 
 & HSNet-HM & \underline{46.7} & \textbf{68.6} & \textbf{71.1} & \textbf{79.7} & \textbf{66.5}  & \underline{53.7}  & \textbf{70.7}  & \textbf{75.2}  & \textbf{83.9} & \textbf{70.9}  \\
 \bottomrule
\end{tabular}
}

\end{table}

% \begin{figure}[t]
% \centering
% \fbox{\includegraphics[height=5.4cm]{figures/qualitave_figure2.png}}
% \vspace{6pt}
% \caption{ \textbf{Visual comparison with HSNet~\cite{HSNet} under several challenging segmentation instances from COCO-20$^i$~\cite{lin2015microsoft} dataset.} Compared to HSNet, the proposed approach produces more accurate segmentation masks that also recover fine-grained details under various appearance variations and complex backgrounds.  \textcolor{green}{need to include other models ?}}

% \label{fig:qualitave_result}
% \end{figure}

\subsubsection{FSS-1000.} 
Table~\ref{table:performance_fss} compares HSNet-HM, VAT-HM, and competing methods on the FSS-1000 dataset~\cite{FSS1000}. In the 1-shot test, HSNet-HM yields a gain of 1.6$\%$ and 1.3$\%$ in mIoU over \cite{HSNet} with ResNet50 and ResNet101 backbones, respectively. In the 5-shot test, we observe an improvement of 0.2$\%$ in mIoU over \cite{HSNet} with the ResNet50 backbone. In the 1-shot test, VAT-HM shows slightly inferior mIoU compared to VAT~\cite{VAT} with ResNet50 but it performs a little better than VAT~\cite{VAT} with ResNet101.

\subsubsection{Generalization test.} 
Following previous works~\cite{HSNet,RePRI}, we perform a domain shift test to evaluate the generalization capability of the proposed method. We trained HSNet-HM and VAT-HM on the COCO-20$^i$ dataset and tested this model on the PASCAL-5$^i$ dataset. The training/testing folds were constructed following ~\cite{RePRI,HSNet}. The objects in training classes do not overlap with the object in the testing classes. As shown in Table~\ref{table:performance_pascal_shift}, HSNet-HM outperforms the current state-of-the-art approaches under both 1-shot and 5-shot tests. In 1-shot test, it delivers a 2$\%$ mIoU gain over RePRI\cite{RePRI} and a 2.4$\%$  mIoU gain over HSNet\cite{HSNet} with ResNet50 and ResNet101 backbones, respectively.
In the 5-shot test, HSNet-HM outperforms HSNet\cite{HSNet} by 1.0$\%$ and 0.6$\%$ in mIoU with ResNet50 and ResNet101 backbones, respectively.

\subsection{Ablation Study and Analysis}
\noindent\textbf{Comparison of the three different masking approaches.}
We compare all three masking approaches, IM \cite{OSLSM}, FM \cite{Zhang2020SGOneSG}, and the proposed HM after incorporating them into HSNet~\cite{HSNet} and evaluate them on the COCO-20$^i$ dataset~(Table~\ref{table:ablation_study_comparison}). We can see that in both 1-shot and 5-shot tests, the proposed HM approach provides noticeable gains over either individual FM and IM techniques. 

% Particularly, in 1-shot test with ResNet50 backbone, the proposed hybrid masking (HM) delivers a gain of 1.4$\%$ over IM. 

\vspace{-6mm}

\begin{table}[!htp]
\caption{ \small Ablation study of the three different masking methods on COCO-20$^i$~\cite{lin2015microsoft}.}
\label{table:ablation_study_comparison}
\centering
\resizebox{\textwidth}{!}{%
\begin{tabular}{@{}cc|cccccc|cccccc@{}}
\toprule
\multirow{2}{*}{\begin{tabular}[c]{@{}c@{}}Backbone\\ feature\end{tabular}} & \multirow{2}{*}{\begin{tabular}[c]{@{}c@{}}Masking\\ methods\end{tabular}} & \multicolumn{6}{c|}{1-shot} & \multicolumn{6}{c}{5-shot} 
\\
 &  & $20^0$ & $20^1$ & $20^2$ & $20^3$ & mIoU & FB-IoU & $20^0$ & $20^1$ & $20^2$ & $20^3$ & mIoU & FB-IoU \\ \midrule
\multirow{3}{*}{ResNet50~\cite{Resnet}} & HSNet-FM~\cite{HSNet} & 36.3 & 43.1 & 38.7 & 38.7 & 39.2& 68.2 & 43.3 & {51.3} & 48.2 & 45.0 & 46.9 & 70.7  \\
 & HSNet-IM & {39.8} & {45.0} & {46.0} & \underline{43.2} & {43.5} & {70.0} & {43.4} & 50.9 & {49.5} & \textbf{48.0} & {47.6} & {71.7} \\
\cmidrule(l){2-14} 
 & HSNet-HM &  \textbf{41.0}&  \underline{45.7}&  \underline{46.9}& \textbf{43.7} & \underline{44.3} & \underline{70.8} & \textbf{45.3} & \underline{53.1}  & \textbf{52.1} & \underline{47.0}  &  \underline{49.4} & {72.2}  \\ 
 \midrule
\multirow{3}{*}{ResNet101~\cite{Resnet}} & HSNet-FM~\cite{HSNet} & 37.2 & 44.1 & 42.4 & 41.3 & 41.2 & 69.1 & {45.9} & 53.0 & \textbf{51.8} & 47.1 & 49.5 & 72.4 \\
 & HSNet-IM & {41.0} & \underline{48.3} & \underline{47.3} & \underline{44.5} & \underline{45.2} & \underline{70.9} & \textbf{46.6} & \underline{54.5} & \underline{50.4} & \underline{47.7} & \underline{49.8} & \underline{72.7} \\ 
 \cmidrule(l){2-14} 
 & HSNet-HM & \textbf{41.2} & \textbf{50.0} & \textbf{48.8} & \textbf{45.9} & \textbf{46.5} & \textbf{71.5}  & \underline{46.5}  & \textbf{55.2} & \textbf{51.8}  & \textbf{48.9}  & \textbf{50.6} & \textbf{72.9} \\

\bottomrule

\end{tabular}
}
\vspace{-12mm}

\end{table}

\begin{table}
  \centering
    \begin{minipage}{.55\textwidth}
\caption{\small Ablation study of the three different merging methods on COCO-20$^i$}
\label{table:Ablation_study}
\centering
\resizebox{\textwidth}{!}{
\begin{tabular}{@{}cc|ccccc}
\toprule
\multirow{2}{*}{\begin{tabular}[c]{@{}c@{}}Feature\\backbone \end{tabular}} & \multirow{2}{*}{\begin{tabular}[c]{@{}c@{}}Methods\end{tabular}} & \multicolumn{5}{c}{1-shot} 
\\
 &  & $20^0$ & $20^1$ & $20^2$ & $20^3$ & mIoU   \\ \midrule
\multirow{3}{*}{ResNet50} & HSNet-HM(Simple Add.)& {40.0} & 43.5 & 43.4 & 43.2 & 42.5 \\
 & HSNet-HM(Reverse) & 39.4 & 45.2 & 42.3 & 41.6 & 42.1 \\
 & HSNet-HM & \textbf{41.0} & \textbf{45.7} & \textbf{46.9} & \textbf{43.7} & \textbf{44.3} \\
\bottomrule 
\end{tabular}
}
    \end{minipage}
    \begin{minipage}{.35\textwidth}
\centering

\caption{ \small Run-time comparison at inference stage on COCO-20$^i$~\cite{lin2015microsoft}.}
\label{tab:runtime_comparison}
\resizebox{\textwidth}{!}{%
\begin{tabular}{@{}c|c|c}
\toprule
\multirow{2}{*}{\begin{tabular}[c]{@{}c@{}}Inference\\Time \end{tabular}} & \multirow{2}{*}{\begin{tabular}[c]{@{}c@{}}secs/image \\  \end{tabular}} &  
Additional \\ &  & Overhead in \%
\\
\midrule
 HSNet-FM  & 0.27 & -  \\
 HSNet-HM  & 0.34 & 25.9   \\
 %w/ HM & 0.27 & 285  \\
\bottomrule
\end{tabular}}
    \end{minipage}
  \end{table}

% \begin{figure}
% \centering
% \includegraphics[height=7.4cm]{figures/qualitave_figure.png}
% \caption{ Qualitative evaluation on COCO-20$^i$}
% \label{fig:qualitave_result}
% \end{figure}

\begin{figure}[!htp]
\centering
{\includegraphics[height=6.2cm]{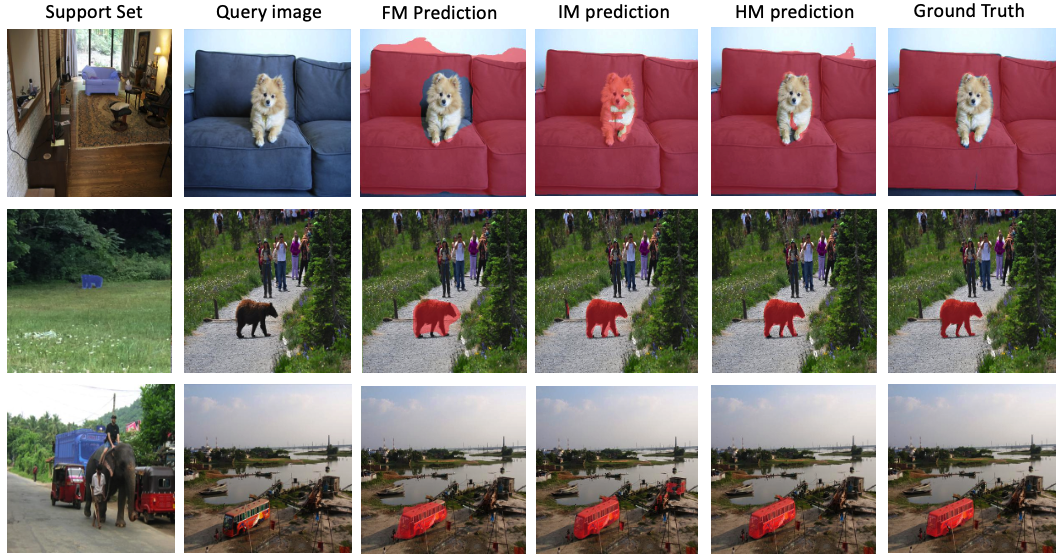}}
\caption{ \small \textbf{Qualitative comparison of three different masking approaches on COCO-$20^i$~\cite{lin2015microsoft} with HSNet.} The blue objects in the support set are the target objects for segmentation. The red pixels are the segmentation results. HSNet-FM can coarsely segment the objects from the background but fails to precisely recover target details, such as target boundaries. HSNet-IM is capable of recovering precise object boundaries, but struggles in distinguishing objects from the background. The proposed approach, HSNet-HM, clearly distinguishes between the target objects and the background and also recovers precise details such as, target boundaries.}
\label{fig:comparison}
\end{figure}

Fig.~\ref{fig:comparison} shows the qualitative results from the three masking methods. The blue objects in the support set are the target objects for segmentation. The red pixels are the segmentation results. FM can coarsely segment the objects from the background but fails to precisely recover target details, such as target boundaries. IM is capable of recovering precise object boundaries, but struggles in distinguishing objects from the background. The proposed approach, HM, clearly distinguishes between the target objects and the background and also recovers precise details such as, target boundaries.

\begin{figure}[!htp]
\centering

{\includegraphics[height=6.1cm]{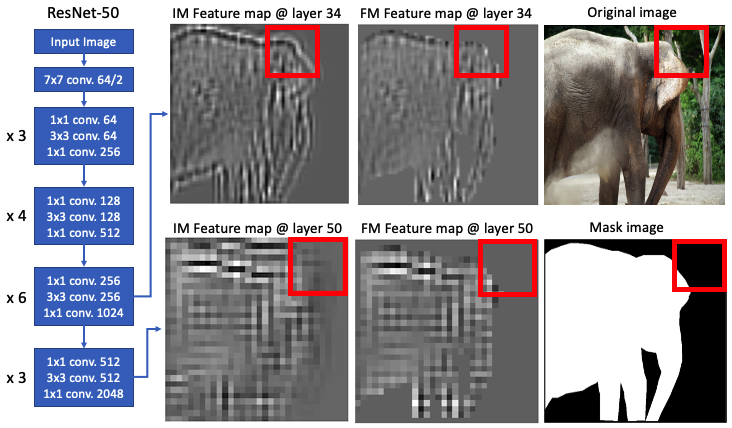}}
\caption{  \textbf{Visual comparison between the feature maps of IM and FM.} These are from ResNet50 at layer34 and layer50. We visualize the first channel of the feature map in grayscale. The feature maps inside the red rectangles reveal that the features from the two feature masking methods are different. Observe activations in the IM feature map at layer 50, it is more indicative of the target object boundaries than the FM feature. Additionally, the IM feature map at layer 34 displays a strong signal around the edge.}% This happens because FM performs masking after getting features. This results in lower accuracy of the boundary of the target object.}
\label{fig:feature_difference}
\end{figure}

Fig.~\ref{fig:feature_difference} shows the visual comparison between the feature maps of IM and FM features. The feature maps inside the red rectangles reveal that the two features produced from the two masking approaches are different. Looking at the area where activations occur in the IM feature map at layer 50, we can see it is more indicative of the target object boundaries than the FM feature. Additionally, looking at the IM feature map at layer 34, we observe that there is a strong signal around the edge and even in side of the target object. This happens because FM performs masking after extracting features, and so this results in less precise target boundaries and loss of texture information. 

\noindent\textbf{Other combination proposals for obtaining HM.} 
We apply various masking methods to create HM features. Various mask sizes are tested by applying dilation to the mask of the support set, but the most plausible result is obtained with the IM masking method. Also, the method in ~\cite{Zhang2020SGOneSG} obtains a mask using the bounding box and applies the average pooling method, but fails to achieve better performance than FM. 
We provide two ablation studies to understand the effectiveness of replacement operation (see Tab.~\ref{table:Ablation_study}).
First, we simply add the corresponding feature maps of IM and FM, denoted as HM(Simple Add). Second, we perform the reverse procedure of the proposed HM. We initialize the HM by IM, and supplement the inactivated features with FM features, denoted as HM(Reverse). Note that, our proposed HM is more effective for FSS compared to HM(Simple Add) and HM(Reverse).

% Initializing the HM by IM, and supplementing inactivate pixels with FM features.
% %
% After initializing HM with IM, inactive pixels are replaced with FM features~(see HM(Reverse) in Tab.~\ref{table:Ablation_study}), and it performs inferior to our HM.

\begin{figure}[!htp]
\centering
\includegraphics[height=3.89cm]{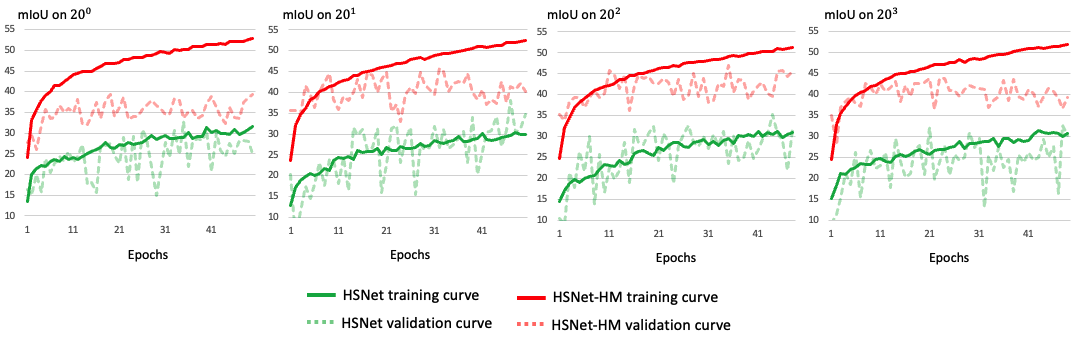}
\caption{ Training profiles of HSNet\cite{HSNet} and HSNet-HM on COCO-20$^i$.}
\label{fig:Training_curve}
\end{figure}

\noindent\textbf{Training efficiency.}
Fig.~\ref{fig:Training_curve} shows the training profiles of HSNet-HM on COCO-20$^i$. 
We see that HM results in faster training convergence compared to HSNet, reducing the training time by a factor of 11x on average. To reach the best model with ResNet101 on COCO-20$^3$, 296.5 epochs are required for HSNet~\cite{HSNet} but HSNet-HM only needs 26.8 epochs on average.
%\hl{VP: how did you judge that?  Val curves in Fig6 are quite noisy, so it would be worth explaining here.  E.g., state (or mark in Fig) how many epochs for HSNet and how many for HSNet+HM.}
%More precisely, HSNet-HM provides approximately 13x faster convergence compared to the HSNet. 
A similar trend was observed in the PASCAL-5$^i$ and FSS-1000 datasets, for which the results are reported in the supplementary material.

\noindent\textbf{Runtime comparison.}
Hybrid masking takes an additional pass over the pixel values to choose between FM and IM. %there may be concerns about the calculation speed.
% costs more than either FM or IM
We measure the computation time of the HM method and other methods for comparison. IM/FM take 0.05 secs/image, and their throughput is 20 images/sec. Whereas HM takes 0.07 secs/image. Therefore, HM induces 40\% more computational time when compared to IM/FM. However, in terms of the model's inference time, our HM adds a relatively less extra overhead~(25.9\%) on top of HSNet~(with FM)~(see Tab.~\ref{tab:runtime_comparison}).

% Although the computation time of the HM method increased by 40$\%$ compared to other methods, resulting in 0.07 images/s, we believe that the $\sim$1300$\%$ improvement in convergence speed is a worthwhile trade-off.
% This high speed was possible due to PyTorch's GPU parallel processing~\cite{pytorch}.

\begin{figure}[!htp]
\centering
\includegraphics[height=4.0cm]{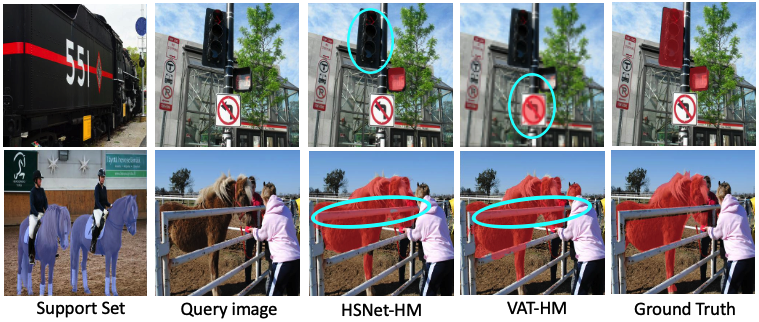}
\caption{ \small Although HSNet-HM/VAT-HM improves mIoU compared to baselines on COCO-20$^i$~\cite{lin2015microsoft}, its performance can be further improved. We identify cases where it struggles to produce accurate segmentation masks~(shown as cyan ellipse).} 
\label{fig:failure_case}
\end{figure}

\noindent \textbf{Limitations.}
% We notice that the performance of proposed HM can be further improved on the PASCAL-5$^i$~\cite{pascal} dataset. 
We found that HSNet-HM performance in PASCAL-5$^i$~\cite{pascal} was inferior to the performance of the COCO-20$^i$~\cite{lin2015microsoft} dataset. A potential reason is that HSNet-HM quickly enters the over-fitting phase due to abundance of information about the target object. The following data augmentation method~\cite{data_aug_1,data_gug_2,VAT} was able to alleviate this problem to some extent, but it did not solve the problem completely. Further, we identify some failure cases for HM~(Fig.~\ref{fig:failure_case}). HM struggles when the target is occluded due to small objects. Also, when the appearance/shape of the target image of the support set and the target image of the query image are radically different.

\vspace{-8mm}
\section{Conclusion}
We proposed a new effective masking approach, termed as hybrid masking. It aims to enhance the feature masking (FM) technique, that is commonly used in existing SOTA methods. We instantiate HM in strong baselines and the results reveal that utilizing HM surpasses the existing SOTA by visible margins and also improves training efficiency. 

\section*{Acknowledgement}
This work was supported in part by NSF IIS Grants \#1955404 and \#1955365.

% Since FM has dominated few-shot segmentation, no one has tried to use IM and no attention has been given to how to modify masking. This simple change to enhance FM significantly improved the current work by a large margin on the most challenging few-shot segmentation benchmark, the COCO-20$^i$ benchmark. This performance result shows that masking should be considered one of the critical factors when designing a network architecture. Furthermore, the fast training convergence will help researchers to test many diverse ideas of new architectures very quickly. 

% Experiment results with four times less learnable parameters show that new network architecture should be investigated not simply making the network complex. 

\bibliographystyle{splncs04}
\bibliography{egbib}

\begin{thebibliography}{10}
\providecommand{\url}[1]{\texttt{#1}}
\providecommand{\urlprefix}{URL }
\providecommand{\doi}[1]{https://doi.org/#1}

\bibitem{RePRI}
Boudiaf, M., Kervadec, H., Masud, Z.I., Piantanida, P., Ben~Ayed, I., Dolz, J.:
  Few-shot segmentation without meta-learning: A good transductive inference is
  all you need? In: Proceedings of the IEEE/CVF Conference on Computer Vision
  and Pattern Recognition (CVPR). pp. 13979--13988 (June 2021)

\bibitem{data_aug_1}
Buslaev, A., Iglovikov, V., Khvedchenya, E., Parinov, A., Druzhinin, M.,
  Kalinin, A.: Albumentations: Fast and flexible image augmentations.
  Information  \textbf{11}, ~125 (02 2020). \doi{10.3390/info11020125}

\bibitem{Deeplab}
Chen, L.C., Papandreou, G., Kokkinos, I., Murphy, K., Yuille, A.L.: Deeplab:
  Semantic image segmentation with deep convolutional nets, atrous convolution,
  and fully connected crfs. IEEE Transactions on Pattern Analysis and Machine
  Intelligence  \textbf{40}(4),  834--848 (2018).
  \doi{10.1109/TPAMI.2017.2699184}

\bibitem{data_gug_2}
Cho, S., Hong, S., Jeon, S., Lee, Y., Sohn, K., Kim, S.: Cats: Cost aggregation
  transformers for visual correspondence. In: Thirty-Fifth Conference on Neural
  Information Processing Systems (2021)

\bibitem{image_net}
Deng, J., Dong, W., Socher, R., Li, L.J., Li, K., Fei-Fei, L.: Imagenet: A
  large-scale hierarchical image database. In: 2009 IEEE Conference on Computer
  Vision and Pattern Recognition (CVPR). pp. 248--255 (2009).
  \doi{10.1109/CVPR.2009.5206848}

\bibitem{pascal}
Everingham, M., Van~Gool, L., Williams, C., Winn, J., Zisserman, A.: The pascal
  visual object classes (voc) challenge. International Journal of Computer
  Vision  \textbf{88},  303--338 (06 2010). \doi{10.1007/s11263-009-0275-4}

\bibitem{pascal_2}
Hariharan, B., Arbel{\'a}ez, P., Girshick, R., Malik, J.: Simultaneous
  detection and segmentation. In: Fleet, D., Pajdla, T., Schiele, B.,
  Tuytelaars, T. (eds.) Computer Vision -- ECCV 2014. pp. 297--312. Springer
  International Publishing, Cham (2014)

\bibitem{Resnet}
He, K., Zhang, X., Ren, S., Sun, J.: Deep residual learning for image
  recognition. In: 2016 {IEEE} Conference on Computer Vision and Pattern
  Recognition (CVPR). pp. 770--778. {IEEE} Computer Society (2016).
  \doi{10.1109/CVPR.2016.90}, \url{https://doi.org/10.1109/CVPR.2016.90}

\bibitem{VAT}
Hong, S., Cho, S., Nam, J., Kim, S.: Cost aggregation is all you need for
  few-shot segmentation. CoRR  (2021)

\bibitem{huang2017densely}
Huang, G., Liu, Z., Van Der~Maaten, L., Weinberger, K.Q.: Densely connected
  convolutional networks. In: Proceedings of the IEEE Conference on Computer
  Vision and Pattern Recognition (CVPR). pp. 4700--4708 (2017)

\bibitem{ASNet}
Kang, D., Cho, M.: Integrative few-shot learning for classification and
  segmentation. In: Proceedings of the {IEEE/CVF} Conference on Computer Vision
  and Pattern Recognition (CVPR) (2022)

\bibitem{Adam}
Kingma, D.P., Ba, J.: Adam: {A} method for stochastic optimization. In: Bengio,
  Y., LeCun, Y. (eds.) 3rd International Conference on Learning
  Representations, {ICLR} 2015, San Diego, CA, USA, May 7-9, 2015, Conference
  Track Proceedings (2015), \url{http://arxiv.org/abs/1412.6980}

\bibitem{krizhevsky2012imagenet}
Krizhevsky, A., Sutskever, I., Hinton, G.E.: Imagenet classification with deep
  convolutional neural networks. Advances in Neural Information Processing
  Systems  \textbf{25} (2012)

\bibitem{ASGNet}
Li, G., Jampani, V., Sevilla-Lara, L., Sun, D., Kim, J., Kim, J.: Adaptive
  prototype learning and allocation for few-shot segmentation. In: CVPR (2021)

\bibitem{FSS1000}
Li, X., Wei, T., Chen, Y.P., Tai, Y.W., Tang, C.K.: Fss-1000: A 1000-class
  dataset for few-shot segmentation. CVPR  (2020)

\bibitem{lin2015microsoft}
Lin, T.Y., Maire, M., Belongie, S., Bourdev, L., Girshick, R., Hays, J.,
  Perona, P., Ramanan, D., Zitnick, C.L., Dollár, P.: Microsoft coco: Common
  objects in context (2015)

\bibitem{liu2016ssd}
Liu, W., Anguelov, D., Erhan, D., Szegedy, C., Reed, S., Fu, C.Y., Berg, A.C.:
  Ssd: Single shot multibox detector. In: European conference on computer
  vision. pp. 21--37. Springer (2016)

\bibitem{FSOT}
Liu, W., Zhang, C., Ding, H., Hung, T.Y., Lin, G.: Few-shot segmentation with
  optimal transport matching and message flow. ArXiv  \textbf{abs/2108.08518}
  (2021)

\bibitem{PPNet}
Liu, Y., Zhang, X., Zhang, S., He, X.: Part-aware prototype network for
  few-shot semantic segmentation (2020)

\bibitem{fully_conv_net_seg}
Long, J., Shelhamer, E., Darrell, T.: Fully convolutional networks for semantic
  segmentation. In: 2015 IEEE Conference on Computer Vision and Pattern
  Recognition (CVPR). pp. 3431--3440. IEEE Computer Society, Los Alamitos, CA,
  USA (jun 2015). \doi{10.1109/CVPR.2015.7298965},
  \url{https://doi.ieeecomputersociety.org/10.1109/CVPR.2015.7298965}

\bibitem{FCN}
Long, J., Shelhamer, E., Darrell, T.: Fully convolutional networks for semantic
  segmentation. In: 2015 IEEE Conference on Computer Vision and Pattern
  Recognition (CVPR). pp. 3431--3440. IEEE Computer Society, Los Alamitos, CA,
  USA (jun 2015). \doi{10.1109/CVPR.2015.7298965},
  \url{https://doi.ieeecomputersociety.org/10.1109/CVPR.2015.7298965}

\bibitem{CWT}
Lu, Z., He, S., Zhu, X., Zhang, L., Song, Y.Z., Xiang, T.: Simpler is better:
  Few-shot semantic segmentation with classifier weight transformer. In: ICCV
  (2021)

\bibitem{receptive_analysis}
Luo, W., Li, Y., Urtasun, R., Zemel, R.: Understanding the effective receptive
  field in deep convolutional neural networks. In: Proceedings of the 30th
  International Conference on Neural Information Processing Systems. p.
  4905–4913. NIPS'16, Curran Associates Inc., Red Hook, NY, USA (2016)

\bibitem{HSNet}
Min, J., Kang, D., Cho, M.: Hypercorrelation squeeze for few-shot segmentation.
  In: Proceedings of the IEEE/CVF International Conference on Computer Vision
  (ICCV). pp. 6941--6952 (October 2021)

\bibitem{FWB}
Nguyen, K., Todorovic, S.: Feature weighting and boosting for few-shot
  segmentation. In: Proceedings of the IEEE/CVF International Conference on
  Computer Vision (ICCV) (October 2019)

\bibitem{Co-FCN}
Rakelly, K., Shelhamer, E., Darrell, T., Efros, A.A., Levine, S.: Conditional
  networks for few-shot semantic segmentation. In: 6th International Conference
  on Learning Representations, {ICLR} 2018, Vancouver, BC, Canada, April 30 -
  May 3, 2018, Workshop Track Proceedings. OpenReview.net (2018),
  \url{https://openreview.net/forum?id=SkMjFKJwG}

\bibitem{redmon2016you}
Redmon, J., Divvala, S., Girshick, R., Farhadi, A.: You only look once:
  Unified, real-time object detection. In: Proceedings of the IEEE conference
  on computer vision and pattern recognition (CVPR). pp. 779--788 (2016)

\bibitem{ren2015faster}
Ren, S., He, K., Girshick, R., Sun, J.: Faster r-cnn: Towards real-time object
  detection with region proposal networks. Advances in neural information
  processing systems  \textbf{28} (2015)

\bibitem{OSLSM}
Shaban, A., Bansal, S., Liu, Z., Essa, I., Boots, B.: One-shot learning for
  semantic segmentation. In: Tae-Kyun~Kim, Stefanos~Zafeiriou, G.B.,
  Mikolajczyk, K. (eds.) Proceedings of the British Machine Vision Conference
  (BMVC). pp. 167.1--167.13. BMVA Press (September 2017).
  \doi{10.5244/C.31.167}, \url{https://dx.doi.org/10.5244/C.31.167}

\bibitem{AMP-2}
Siam, M., Oreshkin, B.N., J{\"{a}}gersand, M.: {AMP:} adaptive masked proxies
  for few-shot segmentation. In: 2019 {IEEE/CVF} International Conference on
  Computer Vision, {ICCV} 2019, Seoul, Korea (South), October 27 - November 2,
  2019. pp. 5248--5257. {IEEE} (2019). \doi{10.1109/ICCV.2019.00535},
  \url{https://doi.org/10.1109/ICCV.2019.00535}

\bibitem{VGG16}
Simonyan, K., Zisserman, A.: Very deep convolutional networks for large-scale
  image recognition. In: Bengio, Y., LeCun, Y. (eds.) 3rd International
  Conference on Learning Representations, {ICLR} 2015, San Diego, CA, USA, May
  7-9, 2015, Conference Track Proceedings (2015),
  \url{http://arxiv.org/abs/1409.1556}

\bibitem{snell2017prototypical}
Snell, J., Swersky, K., Zemel, R.: Prototypical networks for few-shot learning.
  Advances in neural information processing systems  \textbf{30} (2017)

\bibitem{PFENet}
Tian, Z., Zhao, H., Shu, M., Yang, Z., Li, R., Jia, J.: Prior guided feature
  enrichment network for few-shot segmentation. TPAMI  (2020)

\bibitem{episodic}
Vinyals, O., Blundell, C., Lillicrap, T., kavukcuoglu, k., Wierstra, D.:
  Matching networks for one shot learning. In: Lee, D., Sugiyama, M., Luxburg,
  U., Guyon, I., Garnett, R. (eds.) Advances in Neural Information Processing
  Systems. vol.~29. Curran Associates, Inc. (2016),
  \url{https://proceedings.neurips.cc/paper/2016/file/90e1357833654983612fb05e3ec9148c-Paper.pdf}

\bibitem{DAN}
Wang, H., Zhang, X., Hu, Y., Yang, Y., Cao, X., Zhen, X.: Few-shot semantic
  segmentation with democratic attention networks. In: Vedaldi, A., Bischof,
  H., Brox, T., Frahm, J.M. (eds.) Computer Vision -- ECCV 2020. pp. 730--746.
  Springer International Publishing, Cham (2020)

\bibitem{PANet}
Wang, K., Liew, J.H., Zou, Y., Zhou, D., Feng, J.: Panet: Few-shot image
  semantic segmentation with prototype alignment. In: The IEEE International
  Conference on Computer Vision (ICCV) (October 2019)

\bibitem{SAGNN}
Xie, G.S., Liu, J., Xiong, H., Shao, L.: Scale-aware graph neural network for
  few-shot semantic segmentation. In: Proceedings of the IEEE/CVF Conference on
  Computer Vision and Pattern Recognition (CVPR). pp. 5475--5484 (June 2021)

\bibitem{PMM}
Yang, B., Liu, C., Li, B., Jiao, J., Qixiang, Y.: Prototype mixture models for
  few-shot semantic segmentation. In: ECCV (2020)

\bibitem{yang2021mining}
Yang, L., Zhuo, W., Qi, L., Shi, Y., Gao, Y.: Mining latent classes for
  few-shot segmentation. In: ICCV (2021)

\bibitem{CANet}
Zhang, C., Lin, G., Liu, F., Yao, R., Shen, C.: Canet: Class-agnostic
  segmentation networks with iterative refinement and attentive few-shot
  learning. In: Proceedings of the IEEE/CVF Conference on Computer Vision and
  Pattern Recognition (CVPR) (June 2019)

\bibitem{zhang2021fewshot}
Zhang, G., Kang, G., Yang, Y., Wei, Y.: Few-shot segmentation via
  cycle-consistent transformer (2021)

\bibitem{Zhang2020SGOneSG}
Zhang, X., Wei, Y., Yang, Y., Huang, T.: Sg-one: Similarity guidance network
  for one-shot semantic segmentation. IEEE Transactions on Cybernetics
  \textbf{50},  3855--3865 (2020)

\bibitem{pyramidscene}
Zhao, H., Shi, J., Qi, X., Wang, X., Jia, J.: Pyramid scene parsing network.
  In: 2017 IEEE Conference on Computer Vision and Pattern Recognition (CVPR).
  pp. 6230--6239 (2017). \doi{10.1109/CVPR.2017.660}

\end{thebibliography}
\end{document}